\documentclass{article}

\usepackage{microtype}
\usepackage{graphicx}
\usepackage{subfigure}
\usepackage{booktabs} 
\usepackage{xcolor} 
\usepackage{multirow} 
\usepackage{array} 
\usepackage[table,xcdraw]{colortbl}
\usepackage{caption}
\usepackage{stfloats}
\usepackage{algorithm}
\usepackage{algorithmic}
\usepackage{hyperref}

\usepackage[accepted]{icml2025}

\usepackage{amsmath}
\usepackage{amssymb}
\usepackage{mathtools}
\usepackage{amsthm}

\usepackage[capitalize,noabbrev]{cleveref}

\theoremstyle{plain}

\theoremstyle{definition}

\theoremstyle{remark}

\usepackage[textsize=tiny]{todonotes}

\icmltitlerunning{CoS: Chain-of-Shot Prompting for Long Video Understanding}

\begin{document}

\twocolumn[
\icmltitle{CoS: Chain-of-Shot Prompting for Long Video Understanding}




\begin{icmlauthorlist}
\vspace{-5pt}
\icmlauthor{Jian Hu}{QMUL}
\icmlauthor{Zixu Cheng}{QMUL}
\icmlauthor{Chenyang Si}{NTU}
\icmlauthor{Wei Li}{NTU}
\icmlauthor{Shaogang Gong}{QMUL}
\\
\textcolor{purple!80}{https://lwpyh.github.io/CoS/}
\vspace{-5pt}
\end{icmlauthorlist}

\icmlaffiliation{QMUL}{Queen Mary University of London}
\icmlaffiliation{NTU}{Nanyang Technological University}
\icmlcorrespondingauthor{Jian Hu}{jian.hu@qmul.ac.uk}
\vskip 0.3in
]
\printAffiliationsAndNotice{}



\begin{abstract}
Multi-modal Large Language Models (MLLMs) struggle with long
videos due to the need for excessive visual tokens.
These tokens exceed massively the context length of MLLMs,
resulting in filled by redundant task-irrelevant shots.  
How to select shots is an unsolved critical problem: sparse sampling
risks missing key details, while exhaustive sampling overwhelms the
model with irrelevant content, leading to video misunderstanding. 
To solve this problem, we propose \textbf{C}hain-\textbf{o}f-\textbf{S}hot prompting (\textbf{CoS}).
The key idea is to frame shot selection as test-time visual prompt optimisation,
choosing shots adaptive to video understanding semantic task by optimising shots-task alignment.
CoS has two key parts: (1) a binary video summary mechanism that performs pseudo temporal grounding, discovering a binary coding
to identify task-relevant shots, and (2) a video co-reasoning module that deploys the binary coding to pair (learning to align) task-relevant positive shots with irrelevant negative shots. It embeds the optimised shot selections into the original video, facilitating a focus on relevant context to optimize long video understanding.
Experiments across three baselines and five datasets demonstrate the effectiveness and adaptability of CoS.
Code given in \textcolor{purple!80}{\url{https://lwpyh.github.io/CoS}}.

\end{abstract}
\begin{figure}[ht]
   \centering \includegraphics[width=8.5cm]{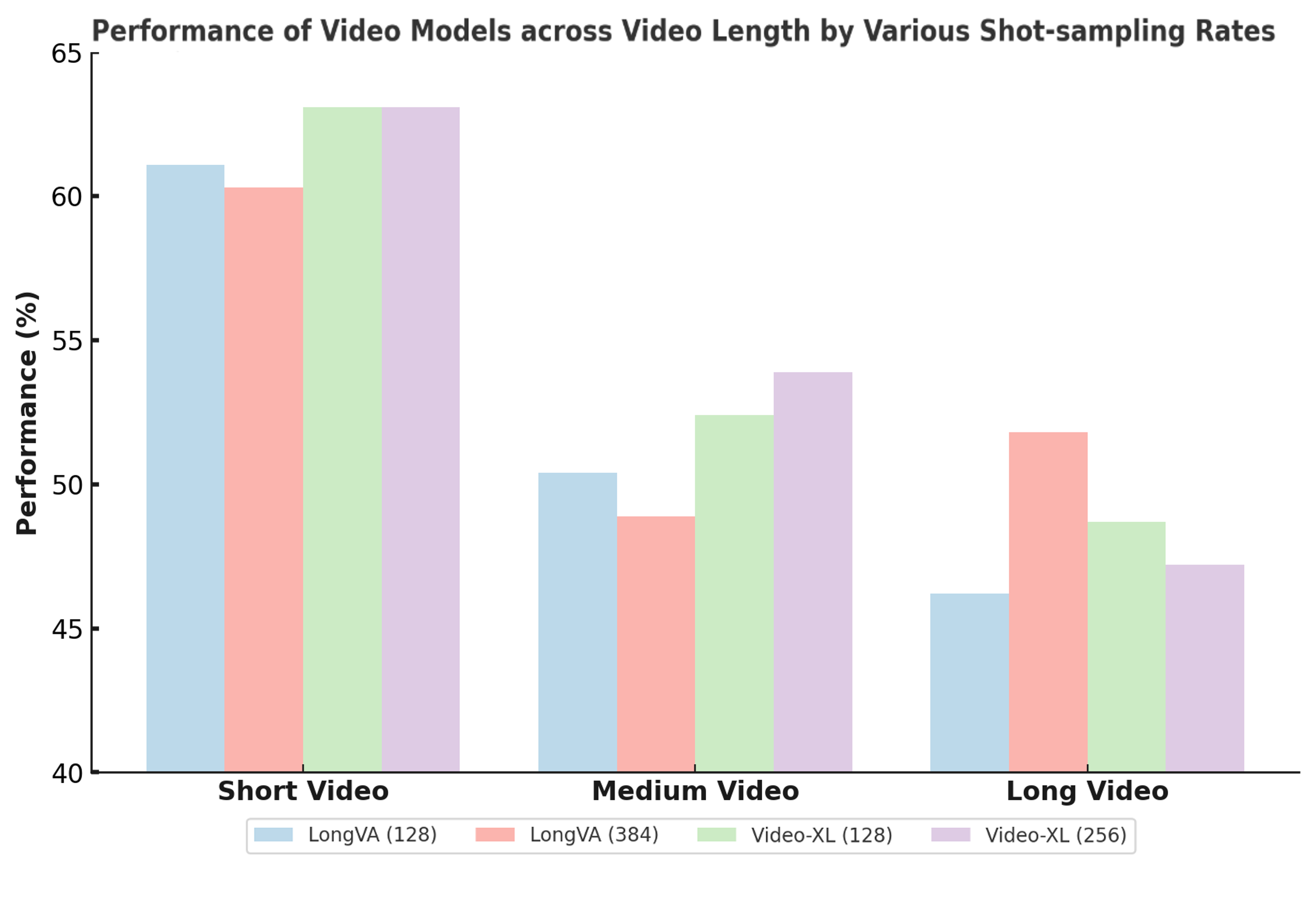}
   \vspace{-30pt}
   \caption{The effects of changing shot-sampling rates on video understanding
     task performance on videos of different lengths in the
     VideoMME~\cite{fu2024video} dataset. Two models are evaluated
     including LongVA~\cite{zhang2024long} and
     Video-XL~\cite{shu2024video}. As the number of sampled shots increased, performance did not consistently improve across various video lengths. That is because while sparse sampling may miss crucial details, exhaustive sampling often overwhelms the model with excessive irrelevant content. This illustrates the key challenge of optimal shot selection especially in long video understanding. That is, how to sample variable details in order to maximise semantic task information extraction whilst minimising distractions from irrelevant details (noise) in video understanding. 
   }\label{fig:motivation}
   \vspace{-10pt}
\end{figure}
\section{Introduction}
\label{Intro}
Driven by advancements in Large Language Models (LLMs)
\cite{chatgpt,jiang2024mixtral,guo2025deepseek}, researchers have
extended LLMs to visual understanding
tasks~\cite{liu2024visual,OpenAIGPT4o,shu2024video}. By modality
alignment and visual instruction tuning, MLLMs have demonstrated
effectiveness in tasks such as captioning and visual question
answering. Despite MLLMs perform well on single images and
short videos (usually under three
minutes)~\cite{zhu2023minigpt,kim2024image}, understanding long videos, such as hour-long videos, remains
a significant problem unsolved~\cite{zhang2024long}. 

\begin{figure*}[ht]
   \centering 
   \vspace{-5pt}
   \includegraphics[width=17.3cm]{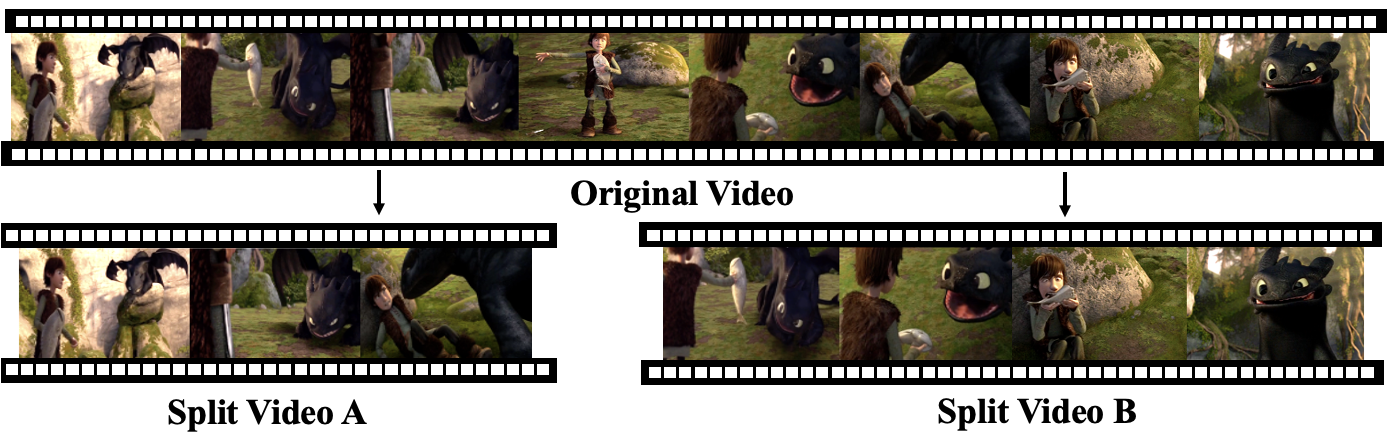}
   \vspace{-25pt}
   \caption{The critical problem of how to select shots in video understanding. 
   In a video that depicts how a boy gradually gains a dragon's trust,
   different sampling methods create two distinct narratives: split
   video A shows the boy being attacked by the dragon, while split
   video B shows him happily sharing food with the dragon. 
   This shows that minor differences in video sampling leads to significant variations in semantic understanding (interpretation).}
   \label{fig:motivation_sec2}
\vspace{-10pt}
\end{figure*}

This challenge arises from the massive visual tokens generated in long videos by contemporary MLLMs, often exceeding the context length and computational capacity of these models, making it computationally intractable.
Existing solutions to extend input capacity include token compression~\cite{li2025llama,wang2024longllava,xue2024longvila} and specialised memory mechanisms~\cite{song2024moviechat,he2024ma,shu2024video}, all aimed at retaining critical information.
However, as shown in Fig.\ref{fig:motivation}, task-relevant shots in
long videos are sparsely distributed. 
Developing an effective sampling strategy is nontrivial and remains an open problem due to two main reasons.
Sampling fewer shots reduces noise and helps the model focuses on relevant information but risks missing critical, sparsely distributed shots. Conversely, sampling more shots captures additional details but introduces significant noise, diluting critical insights. 
In essence, a solution needs not only optimises (minimises)
the number of shots by reducing redundancy and distractions,
but also simultaneously captures (maximises) selectively task-relevant information by reducing omissions. 

Moreover, there is a representation bias problem with existing methods: the role of visual shot
selection in affecting a model’s semantic reasoning process. Current MLLMs mainly process multi-modal inputs by encoding textual and visual information {\em separately}, before cross-modal alignment~\cite{liu2024visual,wang2024qwen2}. 
While input quality can significantly affect performance, most research has focused only on optimising textual prompts for reasoning tasks, neglecting the importance of visual inputs.
For example, VideoCoT~\cite{wang2024videocot} relies on hand-crafted textual prompts, while VoT~\cite{fei2024video1} uses video sense graphs or query decomposition to enhance reasoning. 
Such methods mainly refine text inputs but overlook the optimisation of visual inputs, which is essential for long videos when task-relevant information is sparsely distributed.
As a result, visual selection from the outset (input) becomes critical.
That is illustrated in Fig.\ref{fig:motivation_sec2}, where different shot selections from the same video can lead to entirely different interpretations, demonstrating how video shots can serve as effective visual prompts to guide a model’s reasoning process. However, this is missing in existing methods. 
This oversight highlights an unresolved issue: determining how to optimally sample shots
that can effectively maximise task-relevant information selection whilst simultaneously minimise noise (distractions) in long-video understanding.

In this work, we propose a novel test-time optimisation strategy named
\textbf{C}hain-\textbf{o}f-\textbf{S}hot prompting (\textbf{CoS}). It consists of two parts: Binary Video Summary and
Video Co-Reasoning.
Binary Video Summary identifies sparsely distributed task-relevant
shots by a mosaicing based binary coding on long videos. It leverages MLLMs' reasoning
and summarisation capacity for pseudo temporal
grounding. Video Co-Reasoning then explores this binary coding to construct simultaneously task-relevant positive videos and task-irrelevant negative
videos. This guides the model to focus on critical information while filtering out noise. 
CoS enables test-time model optimisation in long-video understanding by dynamically optimising video inputs during inference. CoS is
training-free and designed for automatic adapting and optimising in task-specific (per video instance) temporal-spatial modelling. Comparative experiments on 17 contemporary models using five
datasets validate the effectiveness of CoS. \textbf{Our contributions are:} 

(1) \textbf{Long-video understanding by visual prompt learning.} We are the first to approach this challenge by optimising input video information to fully utilise the model’s ability to comprehend long videos.
(2) \textbf{Chain-of-Shot prompting (CoS)}, a training-free mosaicing binary coding together with pseudo temporal
grounding is introduced for long video understanding. CoS explores
MLLMs' summary capacity for binary coding and pseudo temporal
grounding on long videos. Moreover, it explores test-time model
optimisation to dynamically construct per video-instance task-specific
positive and negative videos as visual prompts, enabling optimal
selection to capture sparsely distributed task-relevant knowledge in
long videos while minimising interference from irrelevant
information.
(3) \textbf{Comprehensive validation.} Extensive experiments across 5
different datasets on 3 diverse baseline methods against 17 models
demonstrate the effectiveness of CoS.  

\begin{figure*}[ht]
   \centering 
   \includegraphics[width=17.3cm]{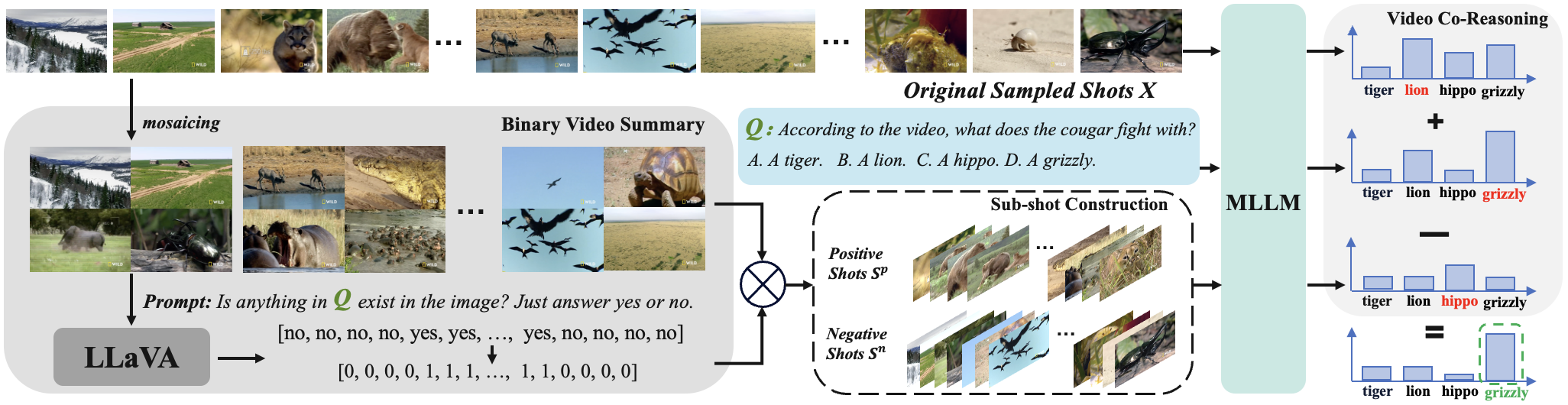}
   \vspace{-15pt}
   \caption{The overall framework of CoS. It first utilises LLaVA to
     perform a mosaicing binary coding to bootstrap video
     summarisation for temporal grounding on a long video. Specifically, every
     four shots are aggregated into a mosaicing composition image. LLaVA
     determines whether task-related elements exist within each
     composition image by encoding a binary value of 1 or 0 (`yes'
     or `no'),
     thereby identifying sparsely distributed task-related shots to
     achieve pseudo temporal grounding. Given this binary video
     summary, task-related positive shots  
$S^p$ and irrelevant negative shots 
$S^n$ are generated and represented by binary codes. 
$S^p$, $S^n$ and the original frame sequence $X$ sampled from  
original video $V$ are then fed into the MLLM for co-reasoning, minimising
     interference of irrelevant video content.
   }\label{fig:framework}
\end{figure*}

\section{Related Works}

\noindent\textbf{MLLMs for visual understanding.} In recent years,
significant progress has been made in the field of MLLMs for visual
understanding~\cite{radford2021learning,zhang2024video,maaz2023video}. Models
like LLaVA~\cite{liu2024visual} achieved cross-modal feature
alignment through projectors, enhancing understanding of single
images. As the focus of research is shifting from image-only models to
those for multi-image and video inputs, various enhancements to the
visual language connector have been proposed. \citet{he2024ma} and
\citet{wang2023chatvideo} implemented average pooling, while
\citet{jin2024chat} and \citet{shu2024video} introduced techniques to dynamically drop
visual tokens. Moreover, \citet{cheng2024videollama} adopted
spatial-temporal convolution to better capture the dynamics of a video
and reduce feature size. 
However, memory constraints and the lack of large-scale annotated
hour-long datasets limit current models. They struggle to process and
understand temporal information in long videos beyond a few minutes,
leading to poor performance on long video understanding. 

\noindent\textbf{MLLMs for Long Video Understanding.} To improve
performance on long videos, several studies have introduced more
fine-grained annotations in datasets at various scales to aid
training~\cite{fu2024video,wu2024longvideobench}. \citet{zhang2024long}
and \citet{he2024ma} extended the context window of LLMs to encompass more
extensive temporal information. LongVILA~\cite{xue2024longvila}
further utilized a parallel processing system to achieve context
compression at the input level. LLaVA-Vid~\cite{li2025llama} and
VideoXL~\cite{shu2024video} sought to obtain a highly compact
representation that preserves key information for effective token
compression. However, these compression techniques invariably lead to
loss of information and poorer video understanding. Critically,
most of these studies focus on learning from the entire video as a single input without selection, neglecting the fact that relevant
information in long videos is often sparsely located. When the presence of irrelevant information is not minimised, it detracts the reasoning power of MLLMs. 

\noindent\textbf{Prompt Engineering.} To enable more effective
reasoning in visual understanding tasks,
VideoCoT~\cite{wang2024videocot} decomposed input questions to
facilitate image-level visual reasoning by MLLMs. Similarly,
VoT~\cite{fei2024video1} used a sense graph and problem decomposition
to enhance short video comprehension and reasoning.  
AoTD~\cite{shi2024unlocking} realized the reasoning of thought chain through agent-of-thought.
VideoGen~\cite{zheng2024videogen} utilised chain-of-thought to assist the video generation process.
~\citet{himakunthala2023let} and \citet{han2024videoespresso} built Chain-of-thought from a dataset perspective to help better evaluate the model's video understanding capabilities.
However, these methods mainly focus on optimising text inputs to
improve reasoning, neglecting the significant temporal changes between
adjacent shots in long videos. Blindly inputting an entire long video
for model processing affect the model’s understanding of both the video and the
questions. Our approach is the first to explore temporal and spatial
modelling on visual inputs for long video understanding, ensuring the visual data better aligns
with text questions and enhances model reasoning on long videos. 

\section{Methodology}
In this work, we introduce a training-free plug-in mechanism called
Chain-of-Shot prompting (CoS), which dynamically optimises the visual
input at test-time per video instance subject to the given video
understanding task. Specifically, given a
video $V$, a video MLLM samples a sequence of shots $X = \{ x_1,
x_2, x_3, \dots, x_n \}$ containing $n$ shots. CoS leverages the
spatial reasoning and summarisation power of a MLLM to perform
binary coding for pseudo temporal grounding. Based on this binary
coding, task-relevant positive shots and irrelevant negative shots are
constructed. These sub-shots, together with the original raw long video, are
input to the MLLM for co-reasoning, allowing the model to effectively
extract task-relevant information and minimise the negative impact of
irrelevant shots, thereby enhancing its reasoning capabilities. 

\subsection{A Closer Look at MLLM Reasoning}
To elaborate on how CoS works, we first revisit how MLLMs typically perform visual understanding tasks.

Given a video $V$ and a query $P$, a shot sampler first uniformly samples $n$ shots to form the set $X$. A MLLM with parameters $\theta$ generates a response $y$ by auto-regressively sampling from a probability distribution conditioned on $P$, $X$, and previously generated tokens:
\begin{equation}
y_t \sim p_{\theta}(y_t \mid X, P, y_{<t}) \propto \exp (\text{logit}_{\theta}(y_t \mid X, P, y_{<t})),
\end{equation}
where $y_t$ denotes the token at time $t$, and $y_{<t}$ represents a sequence of tokens generated up to time $t-1$.

Despite the advanced capabilities of MLLMs, handling long videos
remains a challenge. Task-relevant shots are often sparsely
located and unknown in advance. Low sampling rates may miss these
critical shots. Conversely, increasing the 
sampling rate introduces irrelevant information, making it harder for
the model to focus on key visual features. Subtle variations in visual
inputs can significantly affect the model’s outputs, making it crucial
to balance sampling efficiency and information relevance. 

\subsection{Binary Video Summary}
To provide the model with effective and clear visual inputs, we need
to perform video temporal grounding based on a given query (task),
identifying which shots are related to the task. However, MLLMs
exhibit poor temporal grounding capabilities~\cite{wang2024qwen2}, especially for long
videos where critical information is sparse, and the volume of
irrelevant information is overwhelming. 

While MLLMs often struggle with direct temporal grounding, they possess strong visual reasoning and summary abilities~\cite{liu2024visual}. To leverage these abilities, we perform indirect key shot localization through a binary video summary. Specifically, the model performs spatial localization for each shot to identify whether task-relevant elements exist. By framing this process as a binary classification task (e.g., answering \textit{``yes''} or \textit{``no''}), we achieve a simplified yet effective way to distinguish between relevant and irrelevant shots.
Given a query-specific prompt $P_s$ (\textit{``Is anything in the keyword list present in the image? Just answer yes or no.''$+$} video-specific question $Q_i$), the model processes each shot $x_i$ of the video and outputs a binary result $o_i$: 
\begin{equation} 
o_i = \text{MLLM}(P_s, x_i),
\label{eq:o}
\end{equation}
shots classified as \textit{``yes''} are labelled as task-relevant (positive), while shots classified as \textit{``no''} are labelled as task-irrelevant (negative). This step enables a binary coding of the video shots, where each shot is tagged as either relevant or irrelevant. Consequently, long videos can be summarised into task-relevant and task-irrelevant segments, forming a binary representation of the visual input.

However, this process has two computational problems:
(1) Due to time complexity, evaluating every shot individually is computationally expensive, particularly when the number of sampled shots $n$ is large. (2) Certain temporal-spatial 
 events span multiple consecutive shots (e.g., dynamic actions like cooking), and analysing single shots may fail to capture these temporal dependencies.

To solve these problems, inspired by the idea of using image gird (aka mosaicing) for visual understanding~\cite{kim2024image}, we extend the binary video summary concept by combining every $k$ consecutive shots into $m$ aggregated mosaicing images for reasoning:
\begin{equation} 
A = { a_1, a_2, a_3, \dots, a_m }, \quad \text{where } m = \frac{n}{k}. 
\end{equation}
Each aggregated image $a_s$, consisting of $k$ shots, is processed as a single unit by MLLM with the same prompt $P_s$ as follows:
\begin{equation} 
o_i = \text{MLLM}(P_s, a_s),
\end{equation}
If MLLM outputs \textit{``yes''}, the corresponding group is classified as task-relevant; otherwise, it is deemed irrelevant. This grouping allows us to reduce computational complexity while preserving temporal information across multiple shots. Here, we set $k$ as 4, and LLaVA~\cite{liu2024visual} as the MLLM, More analysis on the hyper-parameter selection is in Tab.\ref{table:PA}.
We use this binary video summary strategy to encode the long video into task-relevant segments for pseudo temporal grounding.
%

\subsection{Video Co-Reasoning}
In long videos, task-relevant shots are usually sparsely distributed,
making it hard for models to identify critical content among
irrelevant information. 
Therefore, we use LLaVA~\cite{liu2024visual} to generate pseudo grounding labels and further process the video to construct balanced sub-shots, providing structured visual inputs for reasoning.

\subsubsection{Constructing Balanced sub-shots}

%
The original video $V$ is first sampled to obtain a sequence of shots $X = \{x_1, x_2, x_3, \dots, x_n\}$, where $n$ is the total number of sampled shots. Based on the MLLM's output, we classify each shot in $X$ as either task-relevant (``yes") or irrelevant (``no"). Shots labelled as "yes" are included in the positive sub-shot $S^p$, while shots labelled as ``no" are included in the negative sub-shot $S^n$. Specifically, the index set of task-relevant shots $\mathcal{I}_{\text{basic}}$ is defined as:
\begin{align}
\mathcal{I}_{\text{basic}} = \{i \mid \text{MLLM output for shot } x_i = \text{"yes"}\},
\end{align}

\noindent\textbf{Positive Shot $S^p$.} 
Task-relevant shots are often sparsely distributed, and directly sampling based on task relevance may result in too few shots, causing significant imbalance between $S^p$ and $S^n$. To ensure $S^p$ includes only task-relevant shots while maintaining a balanced length relative to the video, we adopt the following strategy:
\begin{small}
\begin{align}
S^p_i = 
\begin{cases} 
x_k, & \text{if } k \in [i+1, n] \text{ and } k \in \mathcal{I}_{\text{basic}} \\ 
x_j, & \text{if } k \text{ not found, } j \in [1, i-1] \text{ and } j \in \mathcal{I}_{\text{basic}} \\ 
\text{X}_i, & \text{if no valid } j \text{ or } k \text{ is found.}
\end{cases},
\end{align}
\end{small}
this ensures $S^p$ contains only frames from
$\mathcal{I}_{\text{basic}}$ by prioritising neighbouring key shots
from $\mathcal{I}_{\text{basic}}$, which maintains $S^p$ 's length
consistent with the original video. If no suitable key shots are
found, we set $S^p$ as the original video $X$. 

\begin{table*}[ht]
\centering
\caption{Experimental results on VideoMME benchmarks, we report results with and without subtitle assistance. † indicates that the results were reproduced using their official weights. The best is in \textbf{bold}.}
\setlength{\tabcolsep}{5pt}
\resizebox{\textwidth}{!}{
\begin{tabular}{c|c|c|cccc|cccc}
\hline
\multicolumn{1}{c|}{\multirow{2}{*}{Models}}  & \multicolumn{1}{c|}{\multirow{2}{*}{Size}} & \multicolumn{1}{c|}{\multirow{2}{*}{shots}} & \multicolumn{4}{c|}{VideoMME w/o sub.} & \multicolumn{4}{c}{VideoMME w/ sub.} \\ \cline{4-11}
 &  & & Short & Medium & Long & Avg & Short & Medium & Long & Avg \\ 
\hline
\multicolumn{10}{c}{Proprietary Models} \\ \hline
GPT-4V~\cite{OpenAIGPT4V} & - & 384 & 70.5 & 55.8 & 53.5 & 59.9 & 73.2 & 59.7 & 56.9 & 63.3 \\
GPT-4o~\cite{OpenAIGPT4o} & - & 384 & 80.0 & 70.3 & 65.3 & 71.9 & 82.8 & 76.6 & 72.1 & 77.2 \\
Gemini-1.5-Pro~\cite{team2024gemini} & - & 0.5 fps & \textbf{81.7} & \textbf{74.3} & \textbf{67.4} &\textbf{75.0} & \textbf{84.5} & \textbf{81.0} & \textbf{77.4} & \textbf{81.3} \\
Claude3-Opus~\cite{Claude3} & - & - & 71.0 & 57.4 & 51.2 & 60.0 & 73.5 & 60.1 & 54.7 & 62.9 \\
\hline
\multicolumn{10}{c}{Open-source MLLMs} \\
\hline
VideoChat2~\cite{li2024mvbench} & 7B & 196 & 48.3 & 37.0 & 33.2 & 39.5 & 52.8 & 39.4 & 39.2 & 43.8 \\ %
VideoLLaVA~\cite{lin2023video} & 7B & 49 & 45.3 & 38.0 & 36.2 & 39.9 & 46.1 & 40.7 & 38.1 & 41.6 \\
Sharegpt4Video~\cite{chen2024sharegpt4video} & 7B & 16 & 48.3 & 36.3 & 35.0 & 39.9 & 53.6 & 39.3 & 37.9 & 43.6 \\
InternVL-Chat-V1.5~\cite{chen2024internvl} & 20B & 10 & 60.2 & 46.4 & 45.6 & 47.8 & 61.7 & 49.1 & 46.6 & 52.4 \\
Video-CCAM~\cite{fei2024video} & 14B & 96 & 62.2 & 50.6 & 46.7 & 53.2 & 66.0 & 56.3 & 49.9 & 57.4 \\
Long-LLaVA~\cite{wang2024longllava} & 7B & 128 & 61.9 & 51.4 & 45.4 & 52.9 & 66.2 & 54.7 & 50.3 & 57.1 \\
VITA~\cite{fu2024vita} & 8x7B & 20 & 64.2 & 53.3 & 47.6 & 55.0 & 67.9 & 55.3 & 49.6 & 57.6 \\
Kangaroo~\cite{liu2024kangaroo} & 8B & 64 & 66.1 & 55.3 & 46.7 & 56.0 & 68.0 & 55.4 & 49.3 & 57.6 \\
LongVILA~\cite{xue2024longvila} & 7B & 256 & 69.0 & 58.3 & 53.0 & 60.1 & 72.9 & 64.9 & 57.4 & 65.1 \\
\hline
LongVA~\cite{zhang2024long} & 7B & 128 & 61.1 & 50.4 & 46.2 & 52.6 & 61.6 & 53.6 & 47.6 & 54.3 \\
\rowcolor{purple!10} LongVA+Ours & 7B & 128 & \textbf{61.6} & \textbf{52.0} & \textbf{46.8} & \textbf{53.5} & \textbf{64.2} & \textbf{54.4} & \textbf{48.5} & \textbf{55.7} \\
\hline
Video-XL†~\cite{shu2024video} & 7B & 128 & 63.1 & 52.4	& 48.7 & 54.7 & 68.3 & 55.7 & 52.1 & 58.7 \\
\rowcolor{purple!10} Video-XL+Ours & 7B & 128 & \textbf{64.1} & \textbf{53.6} & \textbf{49.1} & \textbf{55.6} & \textbf{68.9} & \textbf{57.1} & \textbf{52.3} & \textbf{59.5} \\
\hline
LLaVA-Video~\cite{zhang2024video} & 7B & 64 & 76.1 & 61.8 & 52.1 & 63.3 & 78.0 & 69.3 & 61.8 & 69.7 \\
\rowcolor{purple!10} LLaVA-Video+Ours & 7B & 64 & \textbf{77.2} & \textbf{62.4} & \textbf{53.8} & \textbf{64.4} & \textbf{80.1} & \textbf{69.4} & \textbf{65.1} & \textbf{71.5} \\
\hline
\end{tabular}
}
\label{tab:videomme}
\end{table*}

\noindent\textbf{Negative Shot $S^n$.}
For each shot $x_i$ in $X$, if $i \notin \mathcal{I}_{\text{basic}}$, the shot is directly included in $S^n$. We employ the following replacement strategy to ensure $S^n$ primarily contains task-irrelevant content:
\begin{small}
\begin{align}
S^n_i = 
\begin{cases} 
x_k, & \text{if } k \in [i+1, n] \text{ and } k \notin \mathcal{I}_{\text{basic}} \\ 
x_j, & \text{if } k \text{ not found, } j \in [1, i-1] \text{ and } j \notin \mathcal{I}_{\text{basic}} \\ 
\text{black shot}, & \text{if no valid } j \text{ or } k \text{ is found.}
\end{cases},
\end{align}
\end{small}
this ensures $S^n$ captures irrelevant shots while maintaining the same length as $S^p$ and $X$.

\subsubsection{Co-Reasoning with sub-shots}
After constructing the balanced sub-shots $S^p$, $S^n$, and the sampled sequence $X$ from the original video $V$, we jointly input these components into the model for reasoning. The model combines outputs from $X$, $S^p$, and $S^n$ to produce a final response as follows:
\begin{small}
\begin{equation}
\begin{aligned}
\label{eq:final_eq}
y_t &\propto p_\theta(y_t \mid X, Q, y_{<t}) \cdot \left( \frac{p_\theta(y_t \mid S^p, Q, y_{<t})}{p_\theta(y_t \mid S^n, Q, y_{<t})} \right)^{\alpha} \\
&\sim \text{softmax} \big[ \text{logit}_\theta(y_t \mid X, Q, y_{<t}) \\
&\quad + \alpha \cdot \text{logit}_\theta(y_t \mid S^p, Q, y_{<t}) \\
&\quad - \alpha \cdot \text{logit}_\theta(y_t \mid S^n, Q, y_{<t}) \big],
\end{aligned}
\end{equation}
\end{small}
where $\alpha$ is a weighting parameter to adjust the influence of $S^p$ and $S^n$ during reasoning. $Q$ is a question for the video.

\paragraph{Dynamic Weighting Mechanism.}  
Since $S^p$ and $S^n$ are constructed from mutually exclusive pseudo grounding labels, their confidence levels are linked: accurate identification of shots in $S^p$ implies high accuracy for task-irrelevant shots in $S^n$, and vice versa. 

Intuitively, when $S_p$ contains many shots, it closely resembles the sampled sequence $X$, meaning the gain from the pseudo grounding process is limited. Conversely, when $S_p$ contains fewer shots, it indicates that the task-relevant information in the video is sparsely distributed. In this case, the content in $S_p$ and $S_n$ has a significant impact on the MLLM's reasoning. Here, $\alpha$ should increase to amplify the contributions of $S_p$ and $S_n$.
$\alpha$ is defined as:
\begin{equation}
\alpha = 1 - \frac{|S^p|}{|X|},
\label{eq:final_weight}
\end{equation}
where $|S^p|$ is the number of shots in the positive sub-shot, and $|X|$ is the total number of sampled shots. A smaller ratio $\frac{|S^p|}{|X|}$ reflects stronger shot selection.
This mechanism allows the model to adaptively balance its reliance on $S^p$ and $S^n$. When $\alpha$ is large, $S^p$ and $S^n$ have a greater impact, reflecting high confidence in the pseudo grounding. When $\alpha$ is small, the model relies more on $X$, as $S^p$ offers limited additional information. When $\alpha = 0$, the model ignores $S^p$ and $S^n$, reasoning solely with $X$.

\begin{table}[H]
\centering
\vspace{-10pt}
\footnotesize
\setlength{\tabcolsep}{1pt} 
\caption{Experimental results on MLVU and LongVideoBench benchmarks, "LongVideo." refers to LongVideoBench.}
\begin{tabular}{c|c|c|c|c}
\hline
\multicolumn{1}{c|}{\multirow{1}{*}{Models}}  & \multicolumn{1}{c|}{\multirow{1}{*}{Size}} & \multicolumn{1}{c|}{\multirow{1}{*}{shots}} & MLVU & {\small{LongVideo.}} \\ \cline{4-5}
\hline
\multicolumn{5}{c}{Proprietary Models} \\ \hline
GPT-4V~\cite{OpenAIGPT4V} & - & 384 & 49.2 & 60.7 \\
GPT-4o~\cite{OpenAIGPT4o} & - & 384 & 64.6 & 66.7 \\
Gemini-1.5-Pro~\cite{team2024gemini} & - & 0.5 fps & - & 64.4 \\
\hline
\multicolumn{5}{c}{Open-source MLLMs} \\
\hline
VideoChat2~\cite{li2024mvbench} & 7B & 196 & 47.9 & 39.3 \\ %
VideoLLaVA~\cite{lin2023video} & 7B & 49 & 47.3 & 37.6 \\
Shargpt4Video~\cite{chen2024sharegpt4video} & 7B & 16 & 46.4 & 41.8 \\
Video-CCAM~\cite{fei2024video} & 14B & 96 & 63.1 & - \\
\hline
LongVA~\cite{zhang2024long} & 7B & 128 & 56.3 & 47.8 \\
\rowcolor{purple!10} LongVA+Ours & 7B & 128 & \textbf{58.9} & \textbf{52.8} \\\hline
Video-XL†~\cite{shu2024video} & 7B & 128 & 64.3 & 49.8 \\
\rowcolor{purple!10} Video-XL+Ours & 7B & 128 & \textbf{65.2} & \textbf{50.6} \\\hline
LLaVA-Video~\cite{zhang2024video} & 7B & 64 & 70.8 & 58.2 \\
\rowcolor{purple!10} LLaVA-Video+Ours & 7B & 64 & \textbf{71.4} & \textbf{58.9} \\
\hline
\end{tabular}
\vspace{-10pt}
\label{tab:experimental_results_long}
\end{table}

\section{Experiments}
To evaluate the effectiveness of the proposed method, we conducted
experiments with three various baselines on five datasets across
videos of varying lengths, including the VideoMME~\cite{fu2024video}
dataset, the long-video datasets MLVU~\cite{zhou2024mlvu} and
LongVideoBench~\cite{wu2024longvideobench}, as well as two
short-to-medium video datasets, NEXT-QA~\cite{xiao2021next} and
MVBench~\cite{li2024mvbench} for diversity. 

\subsection{Experimental Setup}

\noindent\textbf{Baselines.}  
To validate the effectiveness of CoS, we integrated CoS into three
contemporary long-video understanding baselines:
LongVA~\cite{zhang2024long}, Video-XL~\cite{shu2024video}, and
LLaVA-Video~\cite{zhang2024llavanextvideo}. To ensure robustness, we
evaluated CoS across five datasets:
\textbf{VideoMME}~\cite{fu2024video}: A large-scale dataset containing
videos of varying lengths (short, medium, long) and diverse scenarios,
ideal for evaluating model performance across different temporal
scales. \textbf{MLVU}~\cite{zhou2024mlvu}: A large-scale long-video
dataset featuring diverse scenes and
tasks. \textbf{LongVideoBench}~\cite{wu2024longvideobench}: A
benchmark designed for tasks requiring precise retrieval and reasoning
over detailed multimodal information within extended
inputs. \textbf{MVBench}~\cite{li2024mvbench}: A benchmark cross
over 20 challenging video understanding tasks, focusing on temporal understanding in
dynamic video tasks. It is particularly suited for evaluating CoS's
image concatenation strategy. \textbf{NEXT-QA}~\cite{xiao2021next}: A
short-video benchmark emphasizing causal and temporal reasoning,
challenging models to understand complex sequences and interactions to
answer related questions accurately. 
Additionally, we compared CoS against state-of-the-art general video
understanding methods and long-video understanding approaches (both
open- and closed-source) to comprehensively demonstrate its
effectiveness. 

\noindent\textbf{Metrics.} All five datasets are evaluated using the accuracy metric, where a higher value indicates better performance.

\begin{table}
    \centering
    \footnotesize
    \setlength{\tabcolsep}{1pt} 
    \caption{Results on NEXT-QA and MVBench.}
    \begin{tabular}{c|c|c|c}
        \hline
        \multicolumn{1}{c|}{\multirow{1}{*}{Models}}  & Size & \small{MVBench} & \small{NEXT-QA} \\ 
        \hline
        \multicolumn{4}{c}{Proprietary Models} \\
        \hline
        GPT-4V~\cite{OpenAIGPT4V} & - & 43.5 & - \\
        GPT-4o~\cite{OpenAIGPT4o} & - & - & 76.0 \\
        \hline
        \multicolumn{4}{c}{Open-source MLLMs} \\
        \hline
        mPLUG-Owl~\cite{ye2023mplug} & 7B & 29.7 & 33.8  \\
        Video-LLaVA~\cite{lin2023video} & 7B & - & 40.2 \\
        VideoChat2~\cite{li2024mvbench} & 7B & 51.9 & 78.6 \\
        TimeChat~\cite{ren2024timechat} & 7B & 38.5 & - \\
        ST-LLM~\cite{liu2025st} & 7B & 54.9 & - \\
        PLLaVA~\cite{xu2024pllava} & 7B & 58.1 & 45.6 \\
        Long-LLaVA~\cite{wang2024longllava} & 7B & 54.6 & - \\
        VideoLLava~\cite{lin2023video} & 7B & 52.5 & 71.1 \\
        \hline
        \rowcolor{purple!10}LongVA~\cite{zhang2024long} & 7B & 49.7 & 69.3 \\
        LongVA+Ours & 7B &  \textbf{50.9} & \textbf{69.9}\\
        \rowcolor{purple!10}LLaVA-Video~\cite{zhang2024video} & 7B & 58.6 & 74.2\\
        LLaVA-Video+Ours & 7B & \textbf{60.5} & \textbf{75.1} \\
        \hline
    \end{tabular} \label{table:results_short}
    \vspace{-15pt}
\end{table}

\begin{table*}[t]
    \centering
    \hspace{-6mm}
    \begin{minipage}[t]{0.9\textwidth}
        \centering
        \renewcommand{\arraystretch}{0.9}
        \captionof{table}{Ablation Study on VideoMME with VideoXL and LLaVA-Video}\label{table:ablation}
        \vspace{2pt}
        \setlength{\tabcolsep}{6.5pt}
        \begin{tabular}{ccccc|cccc|cccc}
            \hline
            \multicolumn{5}{c|}{Method's Variants} & \multicolumn{4}{c|}{VideoXL} & \multicolumn{4}{c}{LLaVA-Video} \\
            \hline
            BVS & OFL & PFL & NFL & DWM & short & medium & long & avg & short & medium & long & avg \\
            \hline
            & \checkmark & \checkmark & \checkmark & \checkmark & 63.1 & 52.4 & 48.7 & 54.7 & 76.1 & 61.8 & 52.1 & 63.3\\
            \checkmark & & \checkmark & \checkmark & \checkmark & 52.3 & 45.6 & 47.2 & 48.4 & 58.8 & 52.4 & 51.6 & 54.3\\
            \checkmark & \checkmark &  & \checkmark & \checkmark &  63.8 & 53.3 & 48.8 & 55.3& 76.8& 61.7& 52.6& 63.7\\
            \checkmark & \checkmark & \checkmark & & \checkmark &  63.5 & 53.2 & 48.6 & 55.2&77.1&61.0& 52.0& 63.4\\
            \checkmark & \checkmark & \checkmark & \checkmark & & 63.4 & 53.3 & 48.5 & 55.1&76.5&61.8& 53.1& 63.9\\
            \rowcolor{purple!10}
            \checkmark & \checkmark & \checkmark & \checkmark & \checkmark & \textbf{64.1} & \textbf{53.6} & \textbf{49.1} & \textbf{55.6} & \textbf{77.2} & \textbf{62.4} & \textbf{53.8} & \textbf{64.4}\\
            \hline
        \end{tabular}
    \end{minipage}
\end{table*}

\begin{table*}[t]
    \centering
    \caption{{Parameter ablation study on VideoMME with LongVA as the baseline.}} \label{table:PA}
    \begin{tabular}
      { @{\hspace{-10pt}}
      c @{\hspace{-1pt}}  c@{\hspace{-1pt}}
      c@{\hspace{0pt}}}
      { (a) Various MLLM for Binary video summary.\label{table:SR1}}&
         { (b) Shot-sampling rate.\label{table:SR2}}&
         {(c) Aggregation shot count.\label{table:SR3}}\\
        {
        \footnotesize
        \setlength{\tabcolsep}{1.5pt}
        \renewcommand{\arraystretch}{1.0}
            \begin{tabular}{c|cccc} 
            \hline
            MLLM & short & medium & long & avg \\
            \hline
            LongVA~\cite{zhang2024long} & 58.8 & 49.6 & 43.2 & 50.4\\
            MinichatGPT~\cite{zhu2023minigpt}  & 60.7 & 51.3 & 44.6 & 52.2\\
            Qwen2~\cite{wang2024qwen2}  & 61.2 &\textbf{52.6} & 45.9 & 53.2\\
            \rowcolor{purple!10}LLaVA1.5~\cite{liu2024visual}  & \textbf{61.6} & 52.0 & \textbf{46.8} & \textbf{53.5}\\
            \hline
            \end{tabular}
        }&
        {
        \footnotesize
        \setlength{\tabcolsep}{1.5pt}
        \renewcommand{\arraystretch}{1.0}
            \begin{tabular}{c|cccc}
            \hline
            shots & short & medium & long & avg \\
            \hline
            64 & 61.1 & 50.2 & 44.9 &52.1\\
            96 & 60.9 & 52.0 & 46.1 & 53.0\\
            \cellcolor{purple!10}128 & 
            \cellcolor{purple!10}\textbf{61.6} & \cellcolor{purple!10}\textbf{52.0} & \cellcolor{purple!10}\textbf{46.8} & \cellcolor{purple!10}\textbf{53.5} \\
            192 & 60.8 & 51.8 & 46.0 & 52.9 \\
            \hline
            \end{tabular}
        }&
        {
        \footnotesize
        \setlength{\tabcolsep}{1.5pt}
        \renewcommand{\arraystretch}{1.0}
            \begin{tabular}{c|ccccc}
            \hline
            k & short & medium & long & avg & time (s)\\
            \hline
            2 & 61.2 & 51.9 & 46.9 & 53.3 & 20.7\\
            \cellcolor{purple!10}4 & \cellcolor{purple!10}\textbf{61.6} & \cellcolor{purple!10}\textbf{52.0} & \cellcolor{purple!10}\textbf{46.8} & \cellcolor{purple!10}\textbf{53.5} & \cellcolor{purple!10}15.7 \\
            8 & 60.8 & 52.0 & 46.3 & 53.0 & 13.6 \\
            16 & 60.3 & 51.1 & 46.1 & 52.4 & 11.9\\
            \hline
            \end{tabular}
        }
    \end{tabular}
\end{table*}
\noindent\textbf{Implementation Details.}  
CoS is a training-free, test-time adaptive plug-in. We followed the
shot sampling setup predefined in the baselines for
evaluation. Specifically, we set the sampling rate to 128 shots for
LongVA and Video-XL, and 64 shots for LLaVA-Video. During the binary
coding phase, every four sampled shots are concatenated to form a
composite shot for input into the model, enabling temporal-spatial
modelling. The binary coding process uses LLaVA1.5-13B as the backbone
MLLM. To ensure computational efficiency, we employed 4-bit
quantization and parallel computation using
\texttt{batch\_decode}. Our method runs efficiently on a single 80G
A100 GPU. 
Although our algorithm introduces an additional sample selection module, its inference time complexity and space complexity remain the same as the baseline, both being $O(n)$. More analysis on time costing is in Tab.\ref{table:SR1}(b).

\begin{figure*}[ht]
 \vspace{5pt}
   \centering \includegraphics[width=17cm]{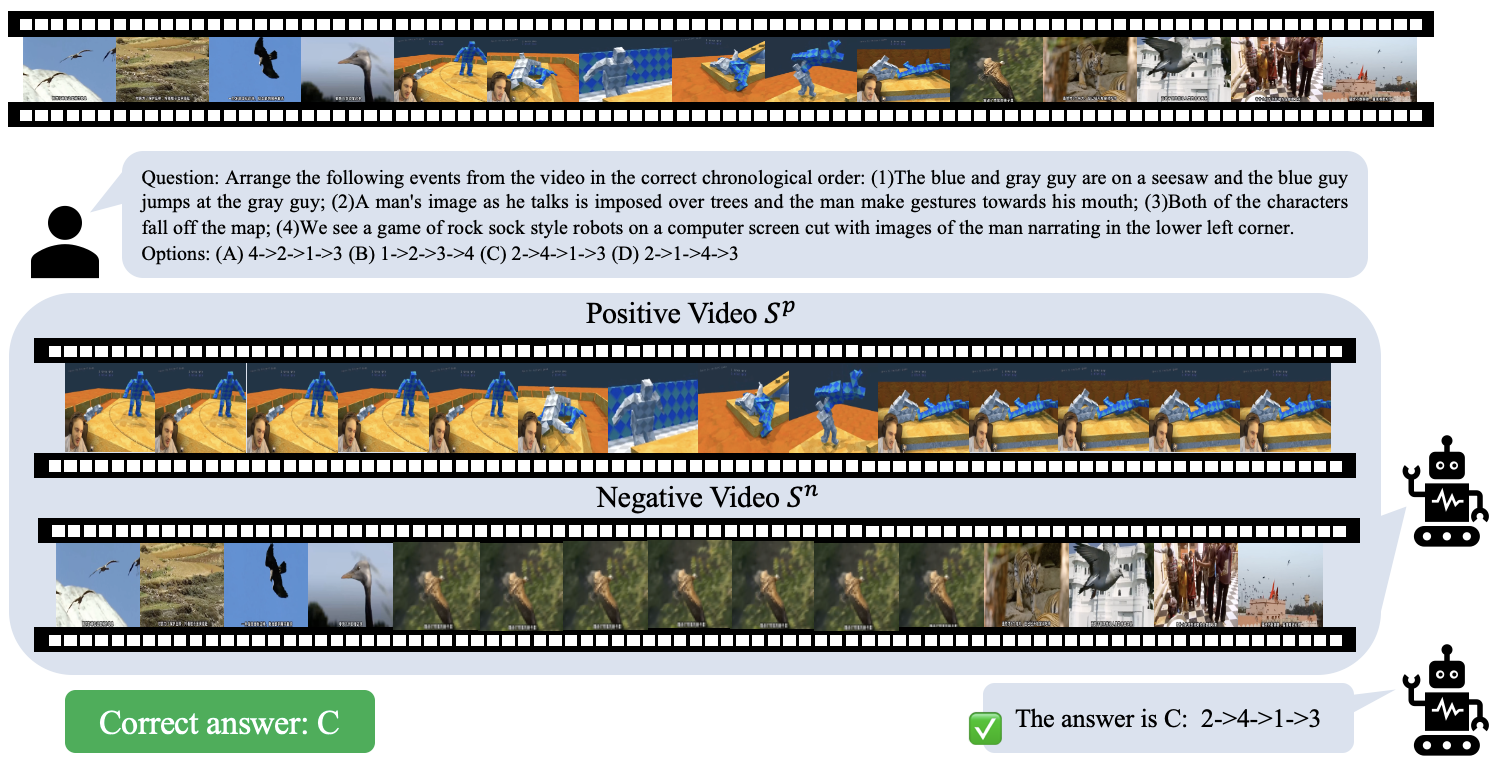}
   \vspace{-10pt}
   \caption{An qualitative evaluation example from MLVU~\cite{zhou2024mlvu} dataset.
   }\label{fig:sample}
\end{figure*}

\subsection{Results and Analysis}
\noindent\textbf{VideoMME.} VideoMME dataset allows for evaluating performance across videos of different lengths. As shown in Tab.\ref{tab:videomme}, we integrated CoS into three baselines and compared the results against closed-source methods and open-source general video methods as well as long-video understanding ones. Evaluations were conducted under both \textit{with subtitle} and \textit{without subtitle} settings. Results show that CoS achieves significant improvements across all baselines and temporal scales (short, medium, and long videos). Notably, CoS exhibits larger performance gains on LLaVA-Video and LongVA, with relatively smaller gains on Video-XL due to its built-in context attention mechanism, which overlaps with CoS’s design. Nevertheless, CoS still delivers improvements, validating its effectiveness.

\noindent\textbf{MLVU and LongVideoBench.}  
These datasets are long video benchmarks. As shown in Tab.\ref{tab:experimental_results_long}, on MLVU's dev set and LongVideoBench's dev set, CoS achieves superior performance compared to all closed-source methods and other open-source 7B-scale models. This demonstrates CoS's strong performance in long-video understanding tasks.

\noindent\textbf{NEXT-QA and MVBench.}  These datasets focus on short-video understanding, including temporal reasoning and inference tasks. As shown in Tab.\ref{table:results_short}, CoS delivers significant improvements on two baselines across both datasets, achieving leading performance on their respective benchmarks. This highlights that CoS's visual prompting modification not only yields gains in long-video tasks but also generalizes well to short-video tasks, underscoring its effectiveness.

\noindent\textbf{Module Analysis.} 
As illustrated in Tab.\ref{table:ablation}, we conducted an ablation
study on the VideoMME dataset using VideoXL and LLaVA-Video as
baselines to assess the impact of various modules. ``BVS" stands for
the binary video summary module, and ``OFL" refers to the original
shots inputted into Eq. \ref{eq:final_eq}, ``PFL" denotes the selected
positive shots inputted into Eq. \ref{eq:final_eq}, and "NFL"
represents the selected negative shots inputted into
Eq. \ref{eq:final_eq}, while ``DWM" is the dynamic weighting
mechanism. In the first row, without the binary video summary module,
only original videos are fed into the MLLM for visual understanding,
causing the model to regress to a baseline model and significantly
underperforming compared to the CoS-enhanced model, thereby
demonstrating the effectiveness of our approach. The second row
removes the original videos in Eq. \ref{eq:final_eq} and relies solely
on the selected positive and negative shots for inference, it falls
significantly compared to CoS, indicating that the content of the
original videos provides a margin of error for the shot selection
strategy. It ensures that incorrectly classified information can still
be processed by the model through the original video feed. The third
and fourth rows evaluate the influence of positive and negative
videos, respectively, indicating that both contribute to visual
understanding. The penultimate row, which omits the dynamic weighting
mechanism, performs worse than the full CoS model, highlighting the
effectiveness of dynamic weighting strategy. 

\noindent\textbf{Shot Selection Model Analysis.} In Tab.\ref{table:PA}(a), we used LongVA~\cite{zhang2024long} as the
baseline to assess the impact of different MLLMs during the binary
video summary phase on shot selection. Initially, we employed LongVA
itself for shot selection, which yields the poorest results. This is because LongVA is better suited for temporal tasks and the entire pipeline relies solely on LongVA for inference, making
it difficult to correct the inherent biases. Performance improves with
other MLLMs such as miniChatGPT~\cite{zhu2023minigpt}, indicating that
employing diverse MLLMs for their respective strengths can better
mitigate the biases a single model might exhibit under unlabelled
conditions. This also suggests that general-purpose MLLMs might
possess superior capabilities in spatial positioning and reasoning
compared to large models specifically designed for videos. The
performance of Qwen2~\cite{wang2024qwen2} and LLaVA1.5 are comparable,
as we leverage visual reasoning and summaries to achieve pseudo
temporal grounding, where Qwen2’s superior temporal reasoning
capabilities have limited scope for impact. 

\noindent\textbf{Impact of Shot-sampling Rate.} As depicted in
Tab.~\ref{table:PA}(b), we utilized LongVA as the baseline to
investigate the impact of different shot-sampling rates on
performance with the VideoMME. When the frame sampling count is
limited to only 64, the performance observed is relatively
mediocre. It is attributed to the inadequate sampling, which fails to
capture essential information effectively. However, as the sampling
count increases, ranging from 96 to 192 frames, the model's
performance exhibits stability, underscoring the robustness of our
approach. It suggests that our CoS is capable of dynamically selecting
the optimal number of shots, thereby efficiently aggregating
information even when the distribution of relevant shots is sparse. 

\noindent\textbf{Image Aggregation Shot Count Analysis.} In
Tab.\ref{table:PA}(c), we used LongVA as the baseline to evaluate the
impact of different image aggregation shot counts on performance with
VideoMME, where "time" indicates the average inference speed per
video. A smaller number of aggregated images results in finer
granularity for pseudo temporal grounding, leading to more accurate
grounding but also increased processing time and difficulty in
capturing temporal relations between shots. Conversely, more
aggregated shots increase the model’s inference speed but reduce the
granularity of pseudo grounding. We find that while an aggregation
count of 2 offers good key shot location ability in longer videos due
to finer grounding granularity, it is more time-consuming. When the
aggregation count exceeds 4, although the inference speed is faster,
the accuracy of pseudo temporal grounding decreases and the increased
number of shots aggregated per image poses challenges in spatial
positioning for the model, leading to a significant decrease in
performance. However, with an aggregation count of 4, the inference
speed is reasonable, and the grounding granularity is moderately
balanced, achieving effective temporal-spatial grounding, hence we
chose an aggregation shot count of 4. 

\noindent\textbf{Qualitative Evaluation.} 
We present qualitative examples of CoS on LLaVA-Video baseline in
Fig.\ref{fig:sample}. CoS$+$LLaVA-Video excels at pinpointing precise
details within extensive videos. This underscores its adeptness at
retrieving and analysing visual data across prolonged
sequences. Moreover, CoS can effectively answer the question by detailing
key characters, settings, and plot events, showcasing its capacity to
handle and interpret exceedingly long videos. 


\vspace{-6pt}
\section{Conclusion} In this work, we introduced a training-free
test-time optimisation plug-in mechanism called Chain-of-Shot
prompting (\textbf{CoS}) for long video understanding. CoS dynamically
selects shots from videos based on per video instance specific query
task, constructing task-relevant positive and task-irrelevant negative
videos from the sparsely distributed useful shots. This approach
enhances models' video understanding ability to comprehend tasks and
achieve better reasoning performance. Extensive experiments
demonstrate the effectiveness of our method. 

\bibliography{example_paper}
\bibliographystyle{icml2025}


\end{document}